\documentclass{article} 
\usepackage{iclr2026_conference,times}

\usepackage{amsmath,amsfonts,bm}

\def\eqref#1{equation~\ref{#1}}

\def\1{\bm{1}}

\DeclareMathAlphabet{\mathsfit}{\encodingdefault}{\sfdefault}{m}{sl}
\SetMathAlphabet{\mathsfit}{bold}{\encodingdefault}{\sfdefault}{bx}{n}

\usepackage{subcaption}
\usepackage{multirow} 
\usepackage{hyperref}
\hypersetup{
  colorlinks=true,
  citecolor=blue,
  linkcolor=blue,
  urlcolor=blue
}
\usepackage{graphicx} 
\usepackage{algorithm}
\usepackage{algpseudocode}
\usepackage{booktabs,tabularx}
\usepackage[most]{tcolorbox}
\tcbuselibrary{listings,breakable}
\usepackage{listings}

\usepackage{booktabs}
\usepackage{url}
\usepackage[table]{xcolor}
\iclrfinalcopy
\usepackage{pifont}
\usepackage[most]{tcolorbox}
\usepackage{enumitem}
\usepackage{array, tabularx}
\newcommand{\xmark}{\ding{55}}  

\usepackage[table]{xcolor} 
\usepackage{tabularx}
\usepackage{booktabs}

\definecolor{cRole}{RGB}{199,21,133}         
\definecolor{cSchema}{RGB}{0,102,204}       
\definecolor{cStyle}{RGB}{0,153,102}        
\definecolor{cPolicy}{RGB}{204,102,0}        
\definecolor{cField}{RGB}{153,0,0}           
\definecolor{cVocab}{RGB}{102,0,153}        
\definecolor{cParts}{RGB}{102,102,0}        
\definecolor{cCoverage}{RGB}{0,128,128}     
\definecolor{cAttr}{RGB}{128,0,64}          
\definecolor{cQuality}{RGB}{0,0,0}          
\definecolor{cFinal}{RGB}{0,90,90}

\providecommand{\cmark}{\checkmark}

\newcommand{\best}[1]{\textbf{#1}}
\newcommand{\second}[1]{\underline{#1}}

\title{Jarvis: Towards Personalized AI Assistant via Personal KV-Cache Retrieval
}

\author{%
\begin{tabular}{c@{\hspace{4em}}c@{\hspace{4em}}c@{\hspace{4em}}c}
{\mdseries\rmfamily Binxiao Xu$^{1,2}$\thanks{Equal contribution. Email: \texttt{a2870578566@gmail.com}}} &
{\mdseries\rmfamily Junyu Feng$^{2}$\footnotemark[1]} &
{\mdseries\rmfamily Shaolin Lu$^{1}$} &
{\mdseries\rmfamily Yulin Luo$^{1}$} \\[1.1ex]
{\mdseries\rmfamily Shilin Yan$^{3}$} &
{\mdseries\rmfamily Hao Liang$^{1}$} &
{\mdseries\rmfamily Ming Lu$^{4}$} &
{\mdseries\rmfamily Wentao Zhang$^{1}$\thanks{Corresponding author. Email: \texttt{wentao.zhang@pku.edu.cn}}}
\end{tabular}
\\[2em]
{\normalfont
$^{1}$Peking University \enspace
$^{2}$Xi'an Jiaotong University \enspace
$^{3}$Alibaba Group \enspace
$^{4}$Intel Labs China
}}

\begin{document}

\maketitle

\begin{abstract}
The rapid development of Vision-language models (VLMs) enables open-ended perception and reasoning. 
Recent works have started to investigate how to adapt general-purpose VLMs into personalized assistants. 
Even commercial models such as ChatGPT now support model personalization by incorporating user-specific information. 
However, existing methods either learn a set of concept tokens or train a VLM to utilize user-specific information. 
However, both pipelines struggle to generate accurate answers as personalized assistants. 
We introduce \textbf{Jarvis}, an innovative framework for a personalized AI assistant through personal KV-Cache retrieval, which stores user-specific information in the KV-Caches of both textual and visual tokens.
The textual tokens are created by summarizing user information into metadata, while the visual tokens are produced by extracting distinct image patches from the user's images.
When answering a question, Jarvis first retrieves related KV-Caches from personal storage and uses them to ensure accuracy in responses.
We also introduce a fine-grained benchmark built with the same distinct image patch mining pipeline, emphasizing accurate question answering based on fine-grained user-specific information. 
Jarvis is capable of providing more accurate responses, particularly when they depend on specific local details. 
Jarvis achieves state-of-the-art results in both visual question answering and text-only tasks across multiple datasets, indicating a practical path toward personalized AI assistants.
The code and dataset will be released.
\end{abstract}

\section{Introduction}

Large vision language models (VLMs) have advanced rapidly in open-ended perception and understanding \citep{internvl25,llavaonevision,qwen2_5_vl,deepseek_v3}. 
However, generating accurate answers based on user-specific information remains challenging, such as consistently recognizing the same pet across different photos and providing accurate answers to questions based on detailed user-specific information.
Two notable failure modes are: (i) models often focus on spurious backgrounds instead of user-specific details, and (ii) many existing methods rely on lengthy prompts, which inflate token budgets and cause latency, instability, and cross-instruction interference. 
Together, these issues lead to inconsistent identity grounding and undermine real-time, user-facing deployment.

We organize LMM personalization along two orthogonal axes. First, should we update parameters per concept or keep the backbone fixed? Second, where is concept information stored at inference time—inside the prompt, within learned tokens/adapters, or in an external cache? Within the parameter-updating branch, prompt- or token-based methods learn lightweight adapters or soft prompts from a few subject images and then use the learned tokens to steer the base model. Yo'LLaVA shows that compact tokens can efficiently capture a subject \citep{Nguyen2024YoLLAVA}. Yo'Chameleon extends this to both understanding and generation under few-shot data while preserving a unified assistant workflow \citep{nguyen2025yo}. UniCTokens goes further by defining a shared pool of learned concept tokens that supports personalized understanding and generation, enabling transfer across tasks without significant architectural changes \citep{An2025UniCTokens}.

\begin{figure}[t]
\centering
\includegraphics[width=\linewidth]{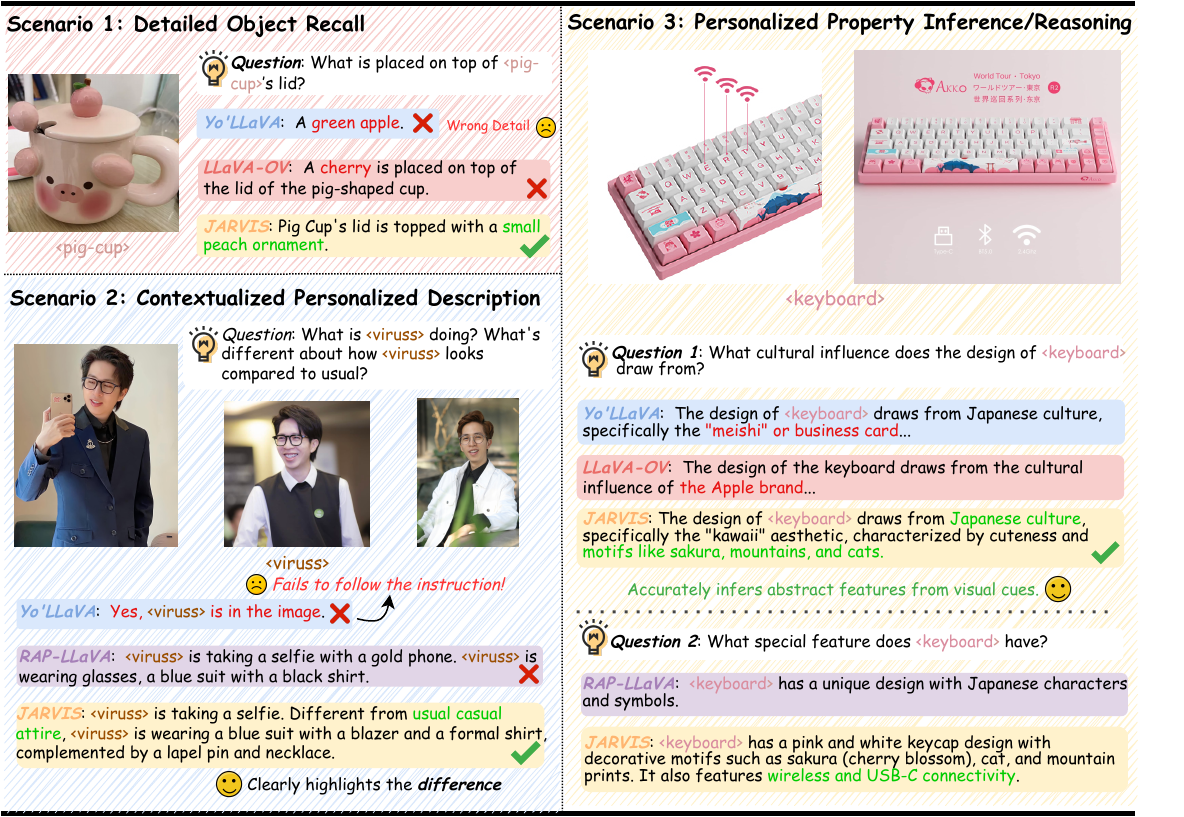}
\caption{\textbf{Qualitative comparison with Yo’LLaVA and RAP-LLaVA across three personalized scenarios.}
(\textbf{1}) \emph{Detailed object recall:} \textbf{Jarvis} correctly recovers fine-grained, instance-specific details (e.g., the small peach ornament) where baselines hallucinate.
(\textbf{2}) \emph{Contextualized personalized description:} \textbf{Jarvis} follows the instruction to contrast with the user’s usual appearance and produces precise, non-generic attributes, while baselines either ignore the comparison or misdescribe.
(\textbf{3}) \emph{Personalized property inference/reasoning:} From visual cues, \textbf{Jarvis} infers abstract cultural aesthetics and functional features (e.g., “kawaii” motifs) with higher faithfulness and completeness.}
\label{fig:demo}
\end{figure}

Complementing the parameter-updating methods above, a parallel line of work reduces dependence on real user data by organizing or synthesizing concept evidence. One strategy is to generate structured surrogates from a handful of seeds. Concept-as-Tree expands each seed into a controlled hierarchy that covers attributes, contexts, and appearance variations, yielding broad concept coverage with low collection cost and reduced privacy risk \citep{an2025concept}. Real-world deployments must also juggle multiple concepts simultaneously; MC-LLaVA combines instruction tuning with personalized prompts to keep multiple concepts distinct yet composable within a single model, avoiding noticeable conflicts \citep{An2025MCLLaVA}. Taken together, these efforts aim to develop a training-free alternative that effectively eliminates the need for per-concept adaptation and the ongoing maintenance burden associated with user-specific adapters or prompts.

Retrieval-centric work removes parameter edits and keeps user evidence external. RAP formalizes a three-stage pipeline: store user knowledge in a key–value store, retrieve it with a multimodal retriever, and condition the model at generation time; this lowers maintenance and enables real-time updates \citep{hao2025rap}. Recent training-free studies further develop this idea by retrieving discriminative fingerprints and reasoning over them without requiring any parameter updates, suggesting that personalization can be both fine-grained and lightweight \citep{Das2025R2P}. Instead of long prompts or broad subject tags, we pursue tighter, concept-level control that yields clear, identity-defining answers and consistency across diverse images.

We adopt a training-free approach, focusing on delivering a seamless real-time user experience. We introduce \textbf{Jarvis}, which converts concept evidence into reusable key--value (KV) states and reuses them across turns without modifying base parameters. Qualitative results in \figurename~\ref{fig:demo} preview Jarvis’s advantages over Yo’LLaVA and RAP-LLaVA across three personalized VQA scenarios—detailed object recall, contextualized personalized description, and property inference. For each concept, we build a concise text profile and extract discriminative visual patches. At inference time, the user query is scored against indexed concept metadata and patch embeddings, and only the top-matching evidence is attached as external KV rather than aggregating all available evidence. We prefetch once and attach on demand, instead of concatenating a long context each turn. This design shortens prompts, lowers latency and compute, and keeps answers grounded to retrieved regions, improving responsiveness and throughput in practical deployments. We implement this design on LLaVA-OneVision \citep{llavaonevision} for both text QA and visual QA. The system covers concept metadata construction, hard patch mining, vector indexing, KV injection, and direct scoring. In addition, we augment the original dataset through the hard patch mining procedure, producing a finer-grained benchmark that more strongly stresses attribute-level grounding and robustness to distractors. This pipeline enables a systematic study of concept understanding and precise grounding under low-latency constraints, while maintaining all base parameters entirely unchanged and ensuring a consistently stable and reproducible execution flow across sessions.

To sum up, our contributions can be concluded as follows:
\begin{itemize}
\item We present Jarvis, a training-free personalization framework that inserts concept evidence via external KV prefill and answers in a single decoding pass without updating base parameters. We instantiate it on LLaVA\mbox{-}OneVision for both text and visual QA.
\item We propose evidence-as-KV caching with a unified text–vision pipeline that compiles compact concept metadata and mines highly discriminative hard patches. The retrieved evidence is attached as an external KV, rather than concatenated prompts, which reduces overall context length and latency while preserving faithful grounding.
\item We release an open-source system and a patch-centric dataset. The package includes code and an end-to-end implementation (segmentation, hard patch mining, indexing, KV injection, direct scoring) plus a patch-guided QA set augmenting Yo'LLaVA and MC\mbox{-}LLaVA.
\end{itemize}

\begin{figure}[t]
\centering
\includegraphics[width=\linewidth]{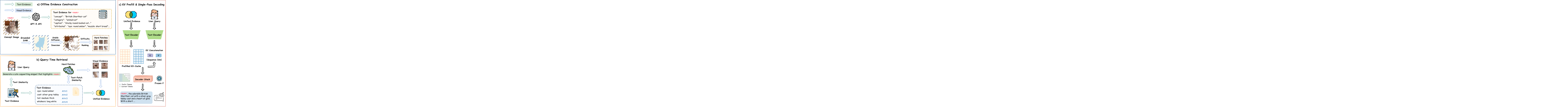}
\caption{\textbf{Model overview.} (\textbf{a}) Offline evidence construction: text metadata synthesis and concept-only hard patch mining. (\textbf{b}) Query-time retrieval: similarity search over text and visual evidence. (\textbf{c}) KV prefill \& single-pass decoding: precompute concept KV states and reuse them during decoding.}
\label{ModelOverview}
\end{figure}

\section{Related Work}

\paragraph{Training-free and retrieval-centric personalization.}
This line of work externalizes user evidence and injects it only at inference time. RAP \citep{hao2025rap} specifies a three-stage pipeline comprising multimodal retrieval, KV-memory storage, and on-the-fly conditioning; this design simplifies maintenance and supports real-time updates. R2P \citep{Das2025R2P} adopts a training-free approach that retrieves discriminative attributes (i.e., concept "fingerprints") and reasons over them without any weight edits. The approach is closely connected to multimodal RAG, which grounds outputs in retrieved evidence~\citep{mei2025survey}. Additionally, it draws on surveys of cross-modal querying and long-horizon user memory~\citep{abootorabi2025ask}, as well as recent progress on discovering subpopulation structure with large language models (LLMs)~\citep{luo2024llm}. Work on agent memory also argues for persisting user state beyond the prompt~\citep{packer2023memgpt}. A practical bottleneck remains: many systems rebuild long prompts at every turn, increasing token usage and latency~\citep{marino2019ok}. A second limitation is granularity: most retrieval pipelines surface coarse descriptions or global profiles rather than fine-grained, attribute- or region-level cues, which reduces fidelity in the presence of distractors and compositional attributes. We address both issues with a training-free alternative that precomputes concept-specific external KV caches reused across turns, populating them with fine-grained text attributes and mined visual patches; this preserves grounding and improves specificity while reducing context length and inference cost.

\section{Method}
\label{sec:method}

We propose \textbf{Jarvis}, a training-free personalization pipeline that injects concept-specific evidence by precomputing and reusing an external key-value (KV) cache. The workflow has three stages aligned with the panels in Fig.~\ref{ModelOverview}: (a) offline evidence construction, (b) query-time retrieval, and (c) KV prefill with single-pass decoding. All base-model parameters remain frozen throughout.

\subsection{Problem Definition}
We study session-level personalization for a single, user-specific concept \(c \in \mathcal{C}\) within a dialogue turn. A concept \(c\) denotes a recurring entity or theme (e.g., a person, pet, or product) that the system resolves at the start of each turn via lightweight retrieval and conditions decoding on the resolved concept to ensure stable grounding and disambiguation across similar contexts.

For each concept we maintain two compact evidence repositories attachable as external key–value (KV) caches rather than prompt tokens: descriptors \(T^{(c)}\) and visual/multimodal patches \(P^{(c)}\). These repositories are curated offline to be small, highly discriminative, and reusable across turns, thereby enabling lower latency and a shorter practical context while preserving fidelity.

\begin{equation}
T^{(c)}=\big\{t_i^{(c)}\big\}_{i=1}^{m_c},\qquad
P^{(c)}=\big\{p_j^{(c)}\big\}_{j=1}^{n_c}.
\end{equation}

We instantiate the global repository \(\mathcal{R}=\{(T^{(c)},P^{(c)})\}_{c\in\mathcal{C}}\) via a compact offline evidence-construction pipeline (Section~\ref{sec:offline-evidence}). Our objective is to produce a response \(y\) that is specific to \(c\), reliably faithful to the query \(q\) (and the image \(I\) when present), and robust against closely visually or semantically similar distractor concepts. Formally, given the resolved concept \(c\),
\begin{equation}
y=\arg\max_{y'} \mathrm{Pr}\!\left(y' \,\middle|\, q,\, I,\, T^{(c)},\, P^{(c)},\, \theta\right).
\label{eq:objective}
\end{equation}

At the beginning of each turn, the system resolves the active concept from the user's explicit mention or through lightweight retrieval-based lookup. Then it selects, attaches, and caches the most relevant evidence as external KV for the current session before decoding the final response.

\begin{figure}[t]
\centering
\includegraphics[width=\linewidth]{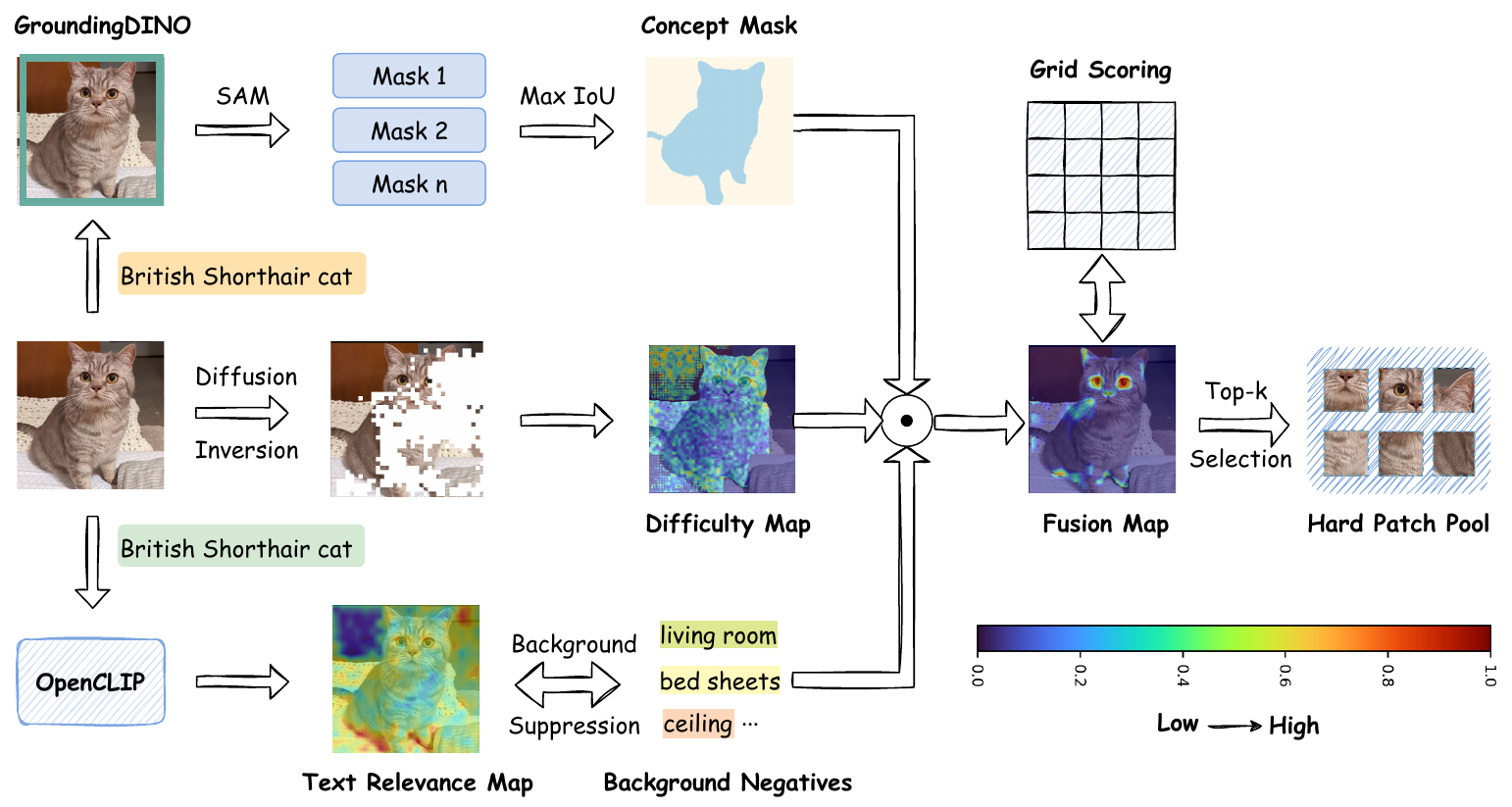}
\caption{\textbf{Concept-only hard patch mining.} We localize the concept, fuse difficulty and text–relevance cues inside the mask, then grid-score to extract top-$k$ informative crops, which populate the visual-indexed \emph{Hard Patch Pool} used by the retriever.}
\label{fig:hard-patch}
\end{figure}

\begin{algorithm}[t]
\caption{Concept-only hard patch mining}
\label{alg:hard-patch}
\begin{algorithmic}[1]
\Require Concept images $\{I_m\}_{m=1}^M$; concept text $T$; background negatives $\mathcal{B}$; grid size $g$; top-$k$; fusion exponent $\gamma$; minimum mask area $\tau$; minimum in-mask coverage $\eta$.
\Ensure Patch set $\mathcal{P}^{(c)}$ with metadata; visual index $\mathcal{I}$.
\Statex

\State $\mathcal{P}^{(c)} \gets \emptyset$;\; $\mathcal{I} \gets \emptyset$;\; $\mathcal{C}_{\text{cand}} \gets [\,]$ \Comment{global candidate list}
\For{$m \gets 1$ \textbf{to} $M$}
  \State \textbf{Subject mask}:\; run GroundingDINO + SAM (Grounded--SAM) on $(I_m, T)$ to get subject boxes; refine with SAM; keep largest connected component (CC) as $M_m$. \textbf{If} $\operatorname{area}(M_m) < \tau$ \textbf{then continue}.
  \State \textbf{Difficulty map}:\; compute $C_m$ via Stable Diffusion inversion (empty prompt); normalize to $[0,1]^{H \times W}$.
  \State \textbf{Text relevance with background suppression}:\; obtain OpenCLIP relevance $R_m^{+}$ for $T$ and background maps $\{R_{m,b}^{-}\}_{b\in\mathcal{B}}$; set
  \State \hspace{1.5em} $R_m \gets \operatorname{normalize}\!\Big(\operatorname{ReLU}\!\big(R_m^{+}-\max_{b\in\mathcal{B}} R_{m,b}^{-}\big)\Big)$
  \State \textbf{Fusion within mask}:\; $C_m^{\mathrm{w}} \gets \operatorname{normalize}\!\big(C_m \cdot (R_m)^{\gamma}\big)\odot M_m$.
  \State \textbf{Grid scoring}:\; tile a fixed $g\times g$ grid on $I_m$.
  \For{\textbf{each} cell $(u,v)$ with box $B_m(u,v)$}
     \State $\kappa \gets \dfrac{|B_m(u,v)\cap M_m|}{|B_m(u,v)|}$;\; \textbf{if} $\kappa < \eta$ \textbf{then continue}.
     \State $s_m(u,v) \gets \operatorname{mean}_{(x,y)\in B_m(u,v)} C_m^{\mathrm{w}}(x,y)$;\; append $\big(m,u,v,s_m(u,v)\big)$ to $\mathcal{C}_{\text{cand}}$.
  \EndFor
\EndFor

\State \textbf{Global selection}:\; choose the top-$k$ elements of $\mathcal{C}_{\text{cand}}$ by $s_m(u,v)$.
\For{\textbf{each} selected $(m,u,v)$}
  \State Crop patch $p$ from $I_m$ at $B_m(u,v)$; record its box, $s_m(u,v)$, and per-cell statistics.
  \State $f(p)\gets \mathrm{CLIP}\text{-}\mathrm{image}\text{-}\mathrm{encoder}(p)$;\; insert $\big(f(p),\text{metadata}\big)$ into index $\mathcal{I}$;\; add $p$ to $\mathcal{P}^{(c)}$.
\EndFor
\State \Return $\mathcal{P}^{(c)},\,\mathcal{I}$
\end{algorithmic}
\end{algorithm}

\subsection{Offline Evidence Construction (Fig.~\ref{ModelOverview}a)}
\label{sec:offline-evidence}
\paragraph{Text metadata.}
We employ the multimodal large language model GPT\mbox{-}5 via the official API \citep{openai2025gpt5} to synthesize a compact textual profile for each concept. Given several representative images and a targeted instruction, the model produces a structured record $\mathcal{T}^{(c)}$ comprising four fields: (i) a canonical name; (ii) a category selected from a fixed taxonomy (e.g., animal, person, device); (iii) an approximately 25\mbox{-}word caption summarizing geometry and salient appearance; and (iv) a list of fingerprint attributes formatted as "part: descriptor" (e.g., eye: round amber). We enforce deterministic decoding (temperature = 0, top-p/top-k disabled) and prescribe a strict key-value JSON schema to ensure format compliance. A lightweight postprocessing step standardizes capitalization and tense, enforces the target schema, and removes invalid or ill-formed entries. The resulting $\mathcal{T}^{(c)}$ is compact, cross\mbox{-}concept consistent, and directly usable for retrieval and KV prefill.

\paragraph{Concept-only hard patches.}
Our goal is to mine patches that enable fine-grained discrimination between concepts across images (Fig.~\ref{fig:hard-patch}; Alg.~\ref{alg:hard-patch}). 
We restrict mining to the subject mask (GroundingDINO+SAM) to avoid background shortcuts and spurious correlations \citep{lin2025perceiveanythingrecognizeexplain}. 
A diffusion–inversion difficulty prior highlights regions that are hard for a generative model to reproduce, which typically coincide with identity-carrying details (e.g., eyes, whisker roots, fine textures). 
In parallel, an OpenCLIP text relevance map focuses on regions aligned with the concept prompt while suppressing common co-occurring backgrounds via negatives, down-weighting cues like blankets or floors that do not define the identity. 
We combine these two signals within the mask to prefer patches that are simultaneously hard to synthesize and semantically on target; an exponent controls the trade-off between the two. 
A fixed $g{\times}g$ grid with an in-mask coverage threshold stabilizes scoring and reduces sensitivity to exact mask boundaries. 
Selecting a global top-$k$ across all images yields a compact pool that is easy to tune and deterministic to reproduce. 
Each selected patch is embedded with the CLIP image encoder and inserted into a visual index, forming the Hard Patch Pool used at retrieval time. 
(Complete implementation details are provided in Alg.~\ref{alg:hard-patch}.)

\subsection{Materializing Evidence as External KV (Fig.~\ref{ModelOverview}c)}
To avoid repeatedly stitching long evidence into the prompt, we convert each concept's text bundle into a short prefix and run a one-time prefill on the frozen base model \(f_{\theta}\). This produces layer-wise, concept-specific key--value (KV) states that are independent of any particular query, which we store externally as reusable evidence units. Because \(\theta\) remains frozen, the upfront prefill cost is amortized across turns and sessions; construction details are provided in Appendix~\ref{app: External KV construction}.

\subsection{Query-Time Retrieval and Answering (Fig.~\ref{ModelOverview}b,c)}
At inference time, the system retrieves a small, relevant concept set from a joint index built over textual bundles and concept-only hard patches, conditioned on the current query and, optionally, an image. The retrieved concepts' cached states are assembled in a fixed, deterministic order and concatenated along the sequence dimension (for both keys and values). The concatenated external prefix is then attached ahead of the current input's cache, allowing decoding to proceed in a single pass. The total external prefix length offsets ensure that relative or rotary position encodings remain consistent after concatenation. We do not introduce head-wise mixing or learnable gating; therefore, the assembly order is fixed for repeatability. When the retrieved set changes across turns, we incrementally prefill only the missing concepts and extend the external cache, avoiding repeated long-prompt construction and reducing effective context length (see Appendix~\ref{app:method} for details).

\section{Experiments}

\subsection{Setup}

\paragraph{Tasks.}
We assess concept personalization under two settings: (i) text-only QA, where the model answers questions about a named concept without an image; and (ii) visual QA (VQA), where questions refer to a held-out image of the concept. In all experiments, the backbone remains frozen.

\paragraph{Compared methods.}
All LLaVA-1.5–derived pipelines are re-implemented on the same LLaVA-OneVision (LLaVA-OV) backbone and vision processor, with shared decoding hyperparameters, prompt templates, and evaluation scripts. We report results on the Yo'LLaVA and MC-LLaVA test sets, and on our fine-grained \texttt{++} variants when applicable.

\begin{itemize}
\item \textbf{Yo'LLaVA} \citep{Nguyen2024YoLLAVA}: single-concept personalization on top of LLaVA. When ported to LLaVA-OV, we reproduce the paper’s single-concept protocol.
\item \textbf{MC-LLaVA} \citep{An2025MCLLaVA}: multi-concept personalization. For a strict apples-to-apples comparison, we evaluate only its single-concept slice, following the authors’ protocol.
\item \textbf{RAP-LLaVA} \citep{hao2025rap}: retrieval-augmented personalization using a concept memory that stores images/attributes and injects retrieved exemplars at inference time. We evaluate under our LLaVA-OV instantiation, consistent with the authors’ settings.
\item \textbf{LLaVA-OV+Prompt}: training-free baseline on LLaVA-OV. We concatenate the query with metadata and concept images into a multi-image context, without caching or retrieval.
\end{itemize}

\paragraph{Datasets.}
We evaluate on \textbf{Yo’LLaVA} \citep{Nguyen2024YoLLAVA} and \textbf{MC-LLaVA} \citep{An2025MCLLaVA}, each organized as personalization episodes with disjoint evidence and evaluation images (i.e., no cross-image leakage). To follow the Yo’LLaVA-data protocol, we report MC-LLaVA results on its single-concept split for a fair, apples-to-apples comparison across settings. In addition, we construct \textbf{Yo’LLaVA++} and \textbf{MC-LLaVA++}: fine-grained, text-only variants guided by patch-centric evidence (hard-patch mining) and GPT-5 generation with light human filtering \citep{openai2025gpt5}, explicitly targeting attribute-level grounding and robustness to distractors. Summary statistics, sampling details, and qualitative examples are provided in Appendix~\ref{app:dataset}.

\paragraph{Protocol overview.}
For each concept, we build a compact text profile from a small set of evidence images and mine candidate hard patches, then precompute a concept-specific external KV cache and index the patches. At inference, we score the user query against the concept attributes and attach only the top-matching textual and visual evidence as external KV. In multi-turn sessions on the same concept, cached \texttt{past\_key\_values} are reused to avoid re-prefill. Unless an ablation states otherwise, hyperparameters are shared across datasets.

\subsection{Main Results}
Table~\ref{tab:main} reports accuracy on \emph{text-only QA} and \emph{VQA} across Yo'LLaVA, MC\mbox{-}LLaVA, and their fine-grained \texttt{++} variants augmented with our hard-patch mining pipeline. Under a unified LLaVA\mbox{-}OV backbone, \textbf{Jarvis} attains the strongest performance in every column while keeping base parameters frozen, across benchmarks and datasets. Beyond the scores, three consistent patterns emerge overall.

\textbf{Fine-grained sensitivity.} Gains are consistently larger on the \texttt{++} splits. These splits emphasize localized, identity-bearing details and suppress background shortcuts. The wider margins indicate that pairing compact text metadata with mined hard patches provides the right inductive bias for disambiguating closely related identities. In contrast, methods that repeatedly inject long textual descriptions at query time raise standard-split scores but yield weaker lift on \texttt{++}, suggesting that fine-grained cues are diluted when evidence is continually concatenated into the prompt.

\textbf{Text-only vs.\ VQA behavior.} \textbf{Jarvis} shows the largest improvements on text-only metrics while also matching or exceeding the strongest VQA baselines, notably in practice and across datasets. This asymmetry is expected: converting concept evidence into external key–value (KV) states supplies a stable semantic prior without inflating the decoder’s working context, which most benefits pure text decoding. Meanwhile, hard patches keep the visual pathway anchored on subject regions, preserving high VQA accuracy without overfitting to co-occurring backgrounds. Prompt-concatenation baselines remain competitive on VQA but leave nontrivial headroom on text-only QA, indicating residual ambiguity when identity must be resolved without an accompanying image.

\textbf{Stability and controllability.} Because concept evidence is prefetched once and then reused across turns, the decoding context stays short and globally consistent. The shorter, stable context reduces prompt interference and yields noticeably steadier behavior across datasets, reflected by uniformly strong results on both Yo'LLaVA and MC\mbox{-}LLaVA. By comparison, systems that rebuild lengthy prompts each turn or interleave retrieval with generation are prone to subtle context drift, which helps explain the mixed performance observed in prior training\mbox{-}free pipelines.

Overall, the results support materializing evidence as external KV and pairing it with concept-only hard patches: the former controls the token footprint. It preserves decoder focus, while the latter supplies localized visual evidence that benefits fine-grained evaluation the most.

\begin{table}[t]
\centering
\caption{\textbf{Performance comparison of personalized VLMs.} “++” denotes our fine-grained split built with the hard-patch mining pipeline. The \textbf{best} and \underline{second-best} numbers are highlighted.}
\label{tab:main}
{\small
\setlength{\tabcolsep}{6pt}%
\renewcommand{\arraystretch}{1.02}%
\resizebox{\linewidth}{!}{%
\begin{tabular}{@{}l c *{2}{c} *{2}{c} c c@{}}
\toprule
\multicolumn{2}{c}{\textbf{Evaluation Dataset}} &
\multicolumn{2}{c}{\textbf{Yo’LLaVA}} &
\multicolumn{2}{c}{\textbf{MC-LLaVA}} &
\multicolumn{1}{c}{\textbf{Yo’LLaVA++}} &
\multicolumn{1}{c}{\textbf{MC-LLaVA++}} \\
\cmidrule(lr){1-2}\cmidrule(lr){3-4}\cmidrule(lr){5-6}\cmidrule(lr){7-7}\cmidrule(lr){8-8}
\multicolumn{1}{c}{\textbf{Method}} & \multicolumn{1}{c}{\textbf{Training-free}} &
VQA & Text-only & VQA & Text-only & Text-only & Text-only \\
\midrule
LLaVA-OV        & \cmark & 0.924 & 0.500 & 0.933 & 0.445 & 0.510 & 0.634 \\
LLaVA-OV+Prompt & \cmark & \second{0.959} & \second{0.823} & \second{0.937} & 0.812 & \second{0.702} & \second{0.679} \\
Yo’LLaVA        & \xmark & 0.929 & 0.800 & 0.655 & 0.658 & 0.663 & 0.646 \\
RAP-LLaVA       & \cmark & 0.917 & 0.795 & 0.844 & \second{0.828} & 0.625 & 0.592 \\
MC-LLaVA        & \xmark & 0.934 & 0.800 & 0.844 & 0.710 & 0.629 & 0.636 \\
\rowcolor{gray!12}
Ours (Jarvis)   & \cmark & \best{0.970} & \best{0.865} & \best{0.941} & \best{0.871} & \best{0.856} & \best{0.835} \\
\bottomrule
\end{tabular}
}
} 
\end{table}

\subsection{Latency and Throughput}\label{sec:lat-qps}
We measure end-to-end responsiveness and serving capacity under identical hardware/decoding: LLaVA\mbox{-}OV backbone and vision tower, greedy decoding, and a fixed max token budget. For each concept, we vary shared queries per session \(Q\in\{1,2,4,8,16,32\}\) and report (i) wall-to-wall \emph{average per-turn latency} (ms, with CUDA sync) and (ii) \emph{throughput} (QPS) as completed requests divided by elapsed wall time at fixed client concurrency. Unless noted, the query is “Tell me \textless mam\textgreater's ear shape, eye color, and hair length.” We compare four pipelines on the same evidence and prompt templates: \emph{Prompt-concat} (user query + compact text metadata + all concept images every turn), \textbf{Jarvis}, Yo’LLaVA, and RAP\mbox{-}LLaVA. The test probes whether Jarvis’s design, which prefetches concept evidence into an external key–value (KV) cache and reuses it across turns, yields lower interactive latency and higher sustained QPS via single-pass decoding with a short, stable context.

\begin{figure}[t]
\centering
\includegraphics[width=\linewidth]{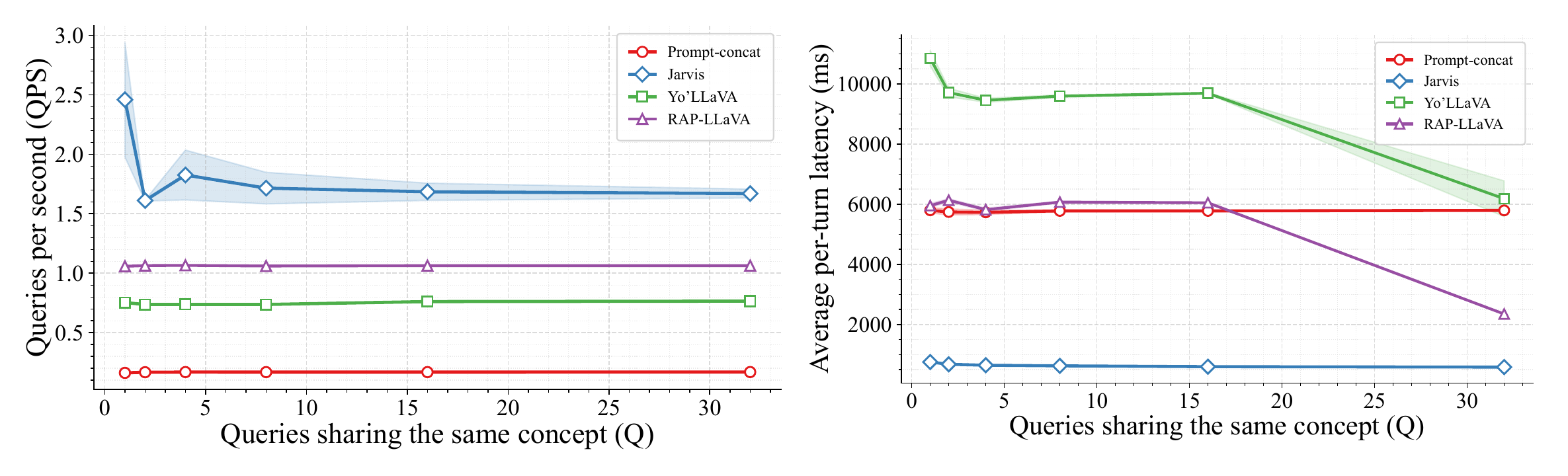}
\caption{Throughput and latency under personalization. 
(Left) QPS vs.\ \(Q\). (Right) average per-turn latency vs.\ \(Q\). 
All methods share hardware/decoding, meaning that 95\% CIs are consistent across trials. 
Higher is better for QPS; lower is better for latency.}
\label{fig:latency-qps}
\end{figure}

Figure~\ref{fig:latency-qps} summarizes results: \textbf{Jarvis} achieves the lowest latency across all \(Q\) and the highest QPS. It avoids rebuilding long prompts and re-encoding multi-image evidence by precomputing a concept-specific KV cache and attaching only top-matching textual attributes plus mined hard patches; decoding then proceeds in one pass on a compact context. In contrast, \emph{Prompt-concat} incurs near-constant per-turn overhead that both elevates latency and caps QPS; Yo’LLaVA and RAP\mbox{-}LLaVA remain competitive but increasingly trail as \(Q\) grows, reflecting the amortization gains of prefilled KV. Concretely, Jarvis reaches \(\sim\!1.6\text{--}2.5\) QPS as \(Q\) increases, whereas Prompt-concat remains at \(\sim\!0.15\text{--}0.18\) QPS—about an order-of-magnitude gap; latency shows the inverse trend (Jarvis lowest, Prompt-concat highest). These trends hold in practice under the same decoding parameters and concurrency, clearly and consistently indicating that materializing concept evidence as external KV reduces token footprint and improves both interactive latency and throughput.

\textbf{Takeaways.} (1) A reusable, prefilled KV turns per-turn prompt construction into a one-time setup, cutting latency and raising QPS as \(Q\) increases. (2) Attaching retrieved top-\(k\) attributes and mined hard patches preserves a short, focused context and yields more stable decoding. (3) For repeated queries about the same concept, \textbf{Jarvis} consistently achieves lower user-perceived latency and higher server-side efficiency than per-turn evidence concatenation.

\begin{table}[t]
\centering
\caption{Ablations of \textbf{Jarvis} components on the same datasets/metrics as Table~\ref{tab:main}. Yo'=Yo'LLaVA, MC=MC-LLaVA, "++" denotes the fine-grained split. VQA denotes VQA accuracy; Txt denotes text-only accuracy. \cmark{}=enabled, \xmark{}=disabled.}
\label{tab:comp-ablate-compact}
\begin{tabular}{ccc cc cc c c}
\toprule
\multicolumn{3}{c}{Components}
 & \multicolumn{2}{c}{Yo’}
 & \multicolumn{2}{c}{MC}
 & Yo’{++} & MC{++} \\
\cmidrule(lr){1-3}\cmidrule(lr){4-5}\cmidrule(lr){6-7}\cmidrule(lr){8-8}\cmidrule(lr){9-9}
QA-Attr & VisPatch & BGS
 & VQA & Txt
 & VQA & Txt
 & Txt & Txt \\
\midrule
\cmark & \cmark & \cmark & \best{0.970} & \best{0.865} & \best{0.941} & \best{0.871} & \best{0.850} & \best{0.835} \\
\cmark & \cmark & \xmark & \second{0.959} & \second{0.850} & 0.936 & \second{0.860} & \second{0.846} & \second{0.824} \\
\cmark & \xmark & \cmark & \best{0.970} & \second{0.855} & \second{0.939} & \second{0.860} & 0.842 & 0.823 \\
\xmark & \cmark & \cmark & 0.935 & 0.703 & 0.924 & 0.662 & 0.627 & 0.573 \\
\bottomrule
\end{tabular}
\end{table}

\subsection{Ablation Studies}
\label{sec:ablation}
We ablate three evidence channels in \textbf{Jarvis} by toggling them on/off—\emph{QA-Attr} (textual attributes drawn from the textual field), \emph{VisPatch} (retrieved hard-patch visual cues), and \emph{BGS} (background suppression that retains only limited local context when forming the hard patch). Results are reported on Yo'LLaVA and MC\mbox{-}LLaVA, together with their fine-grained \texttt{++} splits (Tab.~\ref{tab:comp-ablate-compact}).

\paragraph{QA-Attr (textual attributes).}
Removing QA-Attr yields the most significant degradation across all variants (Tab.~\ref{tab:comp-ablate-compact}). The drop is especially pronounced in text-only evaluation and further amplified on the \texttt{++} splits; VQA also shows a weakening. These trends suggest that textual attributes serve as the dominant semantic prior for identity- and attribute-level personalization. Without this channel, the model drifts toward a non-personalized regime and struggles to resolve fine-grained references.

\paragraph{VisPatch (hard-patch visual evidence).}
Disabling VisPatch leaves VQA on the regular splits nearly unchanged but consistently lowers text-only accuracy and harms the \texttt{++} splits. The effect is stable across both datasets. This aligns with the intuition that localized, hard-to-synthesize visual cues are most valuable when disambiguation hinges on subtle appearance details; by contrast, coarse recognition on standard VQA is already well supported by the textual channel.

\paragraph{BGS (background suppression in patches).}
BGS denotes suppressing the surrounding background when forming the hard patch while retaining limited local context. Enabling BGS yields small, consistent gains, including on the \texttt{++} splits; turning off BGS leads to reproducible drops. Suppressing excess background reduces distractors and retrieval noise, sharpening what is encoded into the cached KV states relative to foreground\mbox{-}only or background\mbox{-}retaining patches.

\paragraph{Cross-variant observations.}
First, the ranking of variants is consistent across datasets: enabling all three channels performs best; dropping BGS or VisPatch yields moderate reductions; dropping QA-Attr causes the sharpest collapse. Second, the \texttt{++} splits are more sensitive than the standard splits, reflecting their reliance on fine-grained cues and localized details. Third, the channels are complementary rather than interchangeable: QA-Attr supplies the semantic prior, VisPatch contributes discriminative visual details, and BGS enhances the robustness of the cached information.

\paragraph{Takeaway.}
Combining QA-Attr, VisPatch, and BGS delivers the strongest and most reliable performance across datasets and metrics. In practice, QA-Attr should be treated as the default semantic scaffold, VisPatch should be enabled whenever fine-grained recognition or identity disambiguation is anticipated, and BGS should be kept to stabilize retrieval and KV construction. This configuration not only maximizes accuracy on the \texttt{++} splits but also preserves competitive VQA on standard splits, offering a balanced recipe that transfers across settings without retuning.

\section{Conclusion}

We introduce a training-free personalization framework that externalizes concept evidence into reusable KV caches and attaches them as a short decoding prefix. This amortizes context processing across turns, lowering token and latency costs while preserving grounding. Compared with light finetuning, external KV improves time-to-first-answer and serving efficiency without per-user adapters. We also enhance the dataset with fine-grained supervision—attribute phrases and region-level patches obtained by automatic mining with lightweight human verification—to raise specificity and better stress-test distractors. Looking ahead, key directions include principled cache composition (such as routing or sparse attention), confidence-aware gating of cache usage, and memory-efficient compression with privacy-preserving storage. Together, these steps retain the throughput gains of \textbf{Jarvis} and improve robustness in open-world deployments.

\section*{Ethics Statement}
All experiments in this study were conducted using publicly available datasets and adhered to the corresponding licenses. No new data collection or clinical trials were performed, and no human subjects research requiring additional consent or IRB approval was involved. The work does not introduce privacy or security risks beyond those inherent to the public datasets, and we declare no conflicts of interest or sponsorships related to this research.

\bibliography{iclr2026_conference}
\bibliographystyle{iclr2026_conference}

\newpage
\appendix

\section{LLM Usage}
Throughout the implementation and debugging process, we consulted large language models for targeted technical guidance. Following the collaborative drafting of the manuscript, we again used LLMs to refine the wording and improve the clarity and overall presentation of the text.

\section{Dataset Statistics}
\label{app:dataset}
We benchmark on \textbf{Yo'LLaVA} and \textbf{MC-LLaVA}, two personalization suites organized as concept-centric episodes with disjoint evidence and evaluation images \citep{Nguyen2024YoLLAVA,An2025MCLLaVA}. To stress fine-grained reasoning without extra training, we also construct text-only variants \textbf{Yo'LLaVA++} and \textbf{MC-LLaVA++} by guiding GPT-5 with mined hard patches and light human filtering \citep{openai2025gpt5}. Here, ``Evid./concept'' denotes the number of images used solely to build evidence; evaluation images never overlap with evidence. Following the \texttt{yollava-data} protocol, we report MC-LLaVA on its single-concept split for comparability. Summary statistics are shown in Table~\ref{tab:datasets}.

\begin{table}[htbp]
\centering
\small
\caption{Datasets used for evaluation. ``Evid./concept'' is the number of images used to construct evidence; evaluation images are disjoint. Yo'LLaVA++ and MC-LLaVA++ are fine-grained variants created via hard-patch guidance and GPT-5 generation with human filtering.}
\label{tab:datasets}
\begin{tabular}{@{}lcccccc@{}}
\toprule
Dataset & Modality & \# Concepts & Evid./Concept & \# QA Pairs & Avg. Q Len. & Split \\
\midrule
Yo'LLaVA    & Text \& Visual & 40  & 5 & 570  & 6.03 & test \\
MC-LLaVA    & Text \& Visual & 118 & 5 & 1055 & 7.16 & test \\
Yo'LLaVA++  & Text-only      & 40  & 5 & 480  & 7.44 & test \\
MC-LLaVA++  & Text-only      & 118 & 5 & 1416 & 8.07 & test \\
\bottomrule
\end{tabular}
\end{table}

\section{Experimental Setup and Hyperparameters}
\label{sec:hyperparameters}

\subsection{Baselines}
For Yo'LLaVA, we adopt the LLaVA-OneVision-Qwen2-7b VLM backbone and train $16$ soft tokens per subject using $5$–$10$ positive images and approximately $150$ hard negatives retrieved per subject. Each subject is taught for up to $15$ epochs, and the best checkpoint is selected based on recognition accuracy on the training split. For RAP-MLLM, we utilize the RAP-LLaVA-13b model provided by the original authors. For each subject, we randomly select one training image as the avatar and sample several images to prompt GPT-5 to generate a description.

\subsection{Evidence Construction}
Unless stated otherwise, each concept uses $k\!=\!5$ evidence images to synthesize a textual profile (canonical name, category, $\sim$25-word caption, and fingerprint attributes) via the GPT-5 API with deterministic decoding (temperature $=0$) \citep{openai2025gpt5}. Hard patches are mined per concept using Algorithm~\ref{alg:hard-patch} with grid size $g\!=\!12$, fusion exponent $\gamma\!=\!1$, background suppression enabled, and top-$k\!=\!4$ patches retained. Text fields are tokenized once to build the concept-specific external KV cache; hard patches populate a compact visual index.

\subsection{Implementation Details}
All methods share the same backbone (LLaVA-OV) and tokenizer. Generation uses greedy decoding unless otherwise noted. KV caches store FP16 tensors and persist per concept. Image resolution, preprocessing, and vision-tower normalization follow the backbone defaults for fairness.

\subsection{Evaluation Protocol}
For each concept: (i) sample $k$ evidence images to build a text profile and mine candidate hard patches; (ii) construct a single text KV cache; and (iii) index all candidate patches. At inference time, embed the query, compute its similarity to all attribute strings, and select the top-$k$ attributes for that turn; the selected attributes and their corresponding top-$k$ visual patches form the evidence fed to the model. In multi-turn sessions on the same concept, cached \texttt{past\_key\_values} are reused to avoid re-prefill.

\section{Methodological Details}
\label{app:method}

\subsection{External KV construction}
\label{app: External KV construction}
Given a concept \(c\) with text bundle \(T^{(c)}\), we linearize it into a short prefix \(\tau^{(c)}\) and run a one-time prefill on the frozen base model \(f_{\theta}\) to obtain layer-wise key–value states,
\begin{equation}
\bigl(\mathbf{K}^{(c)}_{1:L},\,\mathbf{V}^{(c)}_{1:L}\bigr)
= \operatorname{Prefill}\!\bigl(f_{\theta},\,\tau^{(c)}\bigr)\,.
\end{equation}
We store the cache externally as
\[
\mathrm{KV}^{(c)}=\{(\mathbf{K}^{(c)}_{\ell},\mathbf{V}^{(c)}_{\ell})\}_{\ell=1}^{L},
\]
and reuse it across turns; \(\theta\) remains frozen.

\subsection{Retrieval and assembly}
At query time, the system retrieves a small concept set \(\mathcal{S}(q,I)=\{c_1,\dots,c_m\}\).
Let the total external prefix length be
\begin{equation}
L_{\mathrm{ext}}=\sum_{c\in\mathcal{S}(q,I)}\bigl\lvert\tau^{(c)}\bigr\rvert .
\end{equation}
We assemble per-layer external caches by sequence-wise concatenation:
\begin{equation}
\mathbf{K}^{\mathrm{ext}}_{\ell}
=\operatorname{concatseq}\!\big(\mathbf{K}^{(c_1)}_{\ell},\ldots,\mathbf{K}^{(c_m)}_{\ell}\big),\qquad
\mathbf{V}^{\mathrm{ext}}_{\ell}
=\operatorname{concatseq}\!\big(\mathbf{V}^{(c_1)}_{\ell},\ldots,\mathbf{V}^{(c_m)}_{\ell}\big).
\end{equation}

\medskip
\subsection{One-pass decoding}
Given current input \((q,I)\) producing \((\mathbf{K}^{\mathrm{cur}}_{\ell},\mathbf{V}^{\mathrm{cur}}_{\ell})\), we form
\begin{equation}
\tilde{\mathbf{K}}_{\ell}=\operatorname{concatseq}\!\big(\mathbf{K}^{\mathrm{ext}}_{\ell},\mathbf{K}^{\mathrm{cur}}_{\ell}\big),\qquad
\tilde{\mathbf{V}}_{\ell}=\operatorname{concatseq}\!\big(\mathbf{V}^{\mathrm{ext}}_{\ell},\mathbf{V}^{\mathrm{cur}}_{\ell}\big).
\end{equation}
Attention is computed as
\begin{equation}
\mathbf{A}_{\ell}
=\operatorname{softmax}\!\left(\frac{\mathbf{Q}_{\ell}\,\tilde{\mathbf{K}}_{\ell}^{\top}}{\sqrt{d_k}}+\mathcal{M}\right)\tilde{\mathbf{V}}_{\ell}\,,
\end{equation}
where \(\mathbf{Q}_{\ell}\) is the layer-\(\ell\) query projection, \(d_k\) the key dimension, and \(\mathcal{M}\) the attention mask (external prefix fully visible; current tokens autoregressive). Relative/rotary encodings are offset by \(L_{\mathrm{ext}}\) to remain consistent after concatenation. If \(\mathcal{S}(q,I)\) changes across turns, we incrementally prefill only missing concepts and extend the external cache:
\begin{equation}
y=\operatorname{Decode}\!\bigl(
f_{\theta};\ \{\mathrm{KV}^{(c)}\}_{c\in\mathcal{S}(q,I)},\ q,\ I
\bigr)\,.
\end{equation}

\section{Extended Related Work}
\label{app:extended-related}

\subsection{Multimodal LLMs}
Multimodal large language models (LMMs) have advanced general perception and open-ended reasoning by pairing stronger visual encoders with instruction tuning and extensive training on multi-image and video data. Universal embedding objectives further help bridge modality gaps~\citep{gu2025breaking}. In parallel, multi-step visual reasoning with token scaling and verification provides a complementary path toward more reliable inference~\citep{bai2025multi}. Representative systems include LLaVA-OneVision, which unifies single-image, multi-image, and video transfer~\citep{llavaonevision}, InternVL-2.5, which scales to extensive open benchmarks~\citep{internvl25}, and Qwen2.5-VL, which emphasizes precise localization and long-context parsing~\citep{qwen2_5_vl}. Recent foundations consolidate design choices for modern vision language models. MM1 conducts extensive architecture and data ablations~\citep{mckinzie2024mm1}. MM1.5 carries these insights into finetuning~\citep{zhang2025mm1}. Idefics2 offers a practical recipe for grounding and multi-image dialogue~\citep{laurenccon2024matters}. Molmo provides open weights and open data~\citep{deitke2409molmo}. LLaVA-NeXT-Interleave supports multi-image, video, and 3D through interleaved formats~\citep{li2024llava}. Advances in multi-granularity video representation further strengthen the video pathway for multimodal LLMs~\citep{shi2025mavors}. In addition, interleaved document transformation pipelines such as RealSyn improve multimodal processing of mixed-format inputs~\citep{gu2025realsyn}. On the representation-learning side, progress in CLIP-style methods—including ALIP for adaptive language–image pre-training~\citep{Yang_2023_ICCV}, RWKV-CLIP for robust vision–language representation~\citep{gu2024rwkv}, efficient distillation via CLIP-CID~\citep{yang2025clip}, and decoupled global–local alignment for stronger compositional understanding~\citep{hu2025decoupled}—continues to enhance encoder backbones for downstream LMMs. Despite these developments, current models in practice still struggle to represent persistent, user-specific concepts in a reliable way, which motivates the need for explicit personalization mechanisms.

\subsection{Personalization via parameter adaptation}
One research line personalizes models by updating parameters or attaching lightweight modules for each concept. \citet{Nguyen2024YoLLAVA} learn compact subject tokens from a handful of images. \citet{nguyen2025yo} extends from recognition to generation under few-shot constraints. \citet{An2025UniCTokens} unify learned concept tokens to serve understanding and generation in a single interface. \citet{an2025concept} organizes synthetic expansions from a small number of seeds to cover attributes and contexts. \citet{An2025MCLLaVA} address realistic multi-concept composition with instruction tuning and personalized prompts. Beyond these personalized methods, the broader parameter-efficient toolbox reduces per-concept training cost, including LoRA~\citep{hu2022lora}, prefix/prompt tuning~\citep{li2021prefix}, P-Tuning v2~\citep{liu2021p}, and Visual Prompt Tuning~\citep{jia2022visual}, as well as visual-prompt-based instruction for MLLMs~\citep{lin2024draw}. While effective in accuracy, these approaches still require maintaining adapted artifacts per user or concept, which complicates versioning and deployment.

\section{Text Metadata Construction with a Vision--Language API}
\label{app:four-key-metadata}

\subsection{Objective}
We construct, for each concept (object or person), a compact JSON record with exactly four fields: \texttt{concept}, \texttt{category}, \texttt{caption}, and \texttt{fingerprint\_attributes}. The goal is to support downstream retrieval, grounding, and patch-level reasoning with minimal yet highly discriminative metadata.

\subsection{Input organization}
Images are organized under a root directory where each subfolder name serves as a concept identifier. Every subfolder contains one or more photos related to the same concept. Standard formats are accepted (JPEG, PNG, WebP). The subfolder name provides a stable concept hint used to seed the textual fields when the model response omits them.

\subsection{Image preprocessing}
Each image is opened and converted to RGB if necessary. To control bandwidth and ensure consistent visual quality, the long side is constrained to a maximum width of 2048 pixels, while preserving the aspect ratio. The image is then encoded as JPEG with a quality of 92. The encoded bytes are then base64-encoded and embedded into the chat request as data URLs. This keeps the entire interaction self-contained, eliminating the need for external hosting.

\subsection{API request composition}
For each concept, the request payload contains a single textual instruction (the full prompt is provided verbatim later in this subsection) followed by all images of the concept, each attached as an image block via a data URL. The request is issued to a multimodal chat completion endpoint with the following critical settings: JSON-only response enforcement, deterministic decoding, and conservative generation length. Concretely, the generation uses a low temperature (0.0), an explicit request for a JSON object as output, a token limit sufficient to cover the required fields, and a fixed seed. To improve robustness under transient network or service conditions, the client implements bounded exponential backoff with jitter for a small number of retries in response to rate limiting or server errors.

\subsection{Response parsing}
Since models sometimes wrap JSON in explanatory text or code fences, the response is sanitized before parsing. The parser removes code fences if present and scans for the outermost well-formed JSON object. Only that object is then parsed. If no valid object is found, the record for that concept defaults to a minimal placeholder, logging a parsing error for inspection.

\subsection{Schema enforcement and normalization}
The post-processor guarantees that the final record contains exactly the four fields in the expected order. Missing values are imputed as follows: the concept name defaults to the subfolder name when absent; the category and caption default to \texttt{unknown}. The attribute list is normalized to a unique, order-preserving list; if empty, a single placeholder \texttt{``unknown''} is inserted to avoid downstream edge cases. To keep the representation lightweight, the attribute list is capped at a modest upper bound. The output map across all concepts is serialized as human-readable UTF-8 JSON with indentation, preserving non-ASCII characters.

\subsection{Operational details}
Command-line arguments control the image root, output path, model identifier, API base URL, and API key. The API key can be passed via argument or read from an environment variable. Informational logging summarizes progress per concept and reports any parsing or I/O anomalies. Practically, the pipeline tends to be I/O-bound on large folders, while the inference cost scales with image count and attribute richness; the enforced JSON and normalization keep the downstream footprint predictable.

\subsection{Full prompt used for metadata construction}
\noindent The exact instruction sent to the model is reproduced below in verbatim form.

\newcommand{\K}[2]{\textbf{\textcolor{#1}{#2}}} 

\begin{tcolorbox}[
  breakable,
  enhanced,
  title={Four-Key Metadata Prompt (Full)},
  sharp corners=downhill,
  arc=1.5mm,
  boxrule=0.5pt,
  colback=blue!2!white,
  colframe=blue!60!black,
  colbacktitle=blue!10!white,
  coltitle=black,
  borderline west={1.6mm}{0pt}{blue!60!black},
  drop fuzzy shadow,
  left=1.2mm, right=1.2mm, top=0.8mm, bottom=0.8mm
]

\K{cRole}{Role.} You are a vision–language analyst. You will see multiple images of the {\bf same concept} (same object/person). Produce a minimal JSON.

\vspace{2mm}
\K{cSchema}{Output schema (exact keys, exact order).}
\begin{tcolorbox}[colback=white, colframe=black!10, boxrule=0.3pt, left=1mm, right=1mm, top=0.6mm, bottom=0.6mm]
\ttfamily\footnotesize
{"concept": "string", "category": "string", "caption": "string", "fingerprint\_attributes": ["...", "..."]}
\end{tcolorbox}

\K{cStyle}{Global style (English only).} Telegraphic style, commas and hyphenated compounds, lowercase everywhere except exact wordmarks/logos (keep casing), no periods, separate clauses with commas.

\vspace{1mm}
\K{cPolicy}{Evidence policy (silent).} Use visual evidence only. Tally traits per image; include traits present in $\ge 60\%$; prefix {\tt often:} for $40\text{--}60\%$; never include $<40\%$. Resolve conflicts by majority with ties toward more discriminative cues (shape $<$ color/pattern $<$ material $<$ background). Include background only if stable \& discriminative. Allow negative but discriminative cues with {\tt none}/{\tt absent} (e.g., {\tt tail: none}, {\tt logo: absent}). Do not infer brand/material if uncertain. If identity/name is unknown, do not guess.

\vspace{1mm}
\K{cField}{Field rules.}
\begin{itemize}[leftmargin=4.2mm,itemsep=2pt,topsep=2pt]
  \item \K{cField}{concept (3–7 tokens):} signature phrase aiding retrieval \& disambiguation; include $\ge$1 discriminative token (breed/shape/color/role); avoid verbatim reuse of any {\tt [part]: [descriptor]}; examples: {\tt shiba inu plush, curled-tail gray}, {\tt cat ceramic mug}, {\tt male adult with glasses}; if unclear write {\tt unknown}.
  \item \K{cField}{category (1–3 tokens):} short normalized label, open-world; may use simple hierarchy with {\tt <} (e.g., {\tt animal<dog}, {\tt device<keyboard>}); pick the most defensible label; avoid long phrases and overfitting to fine-grained names when uncertain.
  \item \K{cField}{caption (24–30 words):} concept-level majority summary in strict order—(1) silhouette/shape, (2) dominant colors/patterns, (3) two–three signature parts, (4) stable accessory/wordmark/material. Do {\bf not} include {\tt often:}; do {\bf not} copy any exact {\tt [part]: [descriptor]} string; paraphrase at a higher level; optionally append {\tt palette: X dominant; Y secondary}.
  \item \K{cField}{fingerprint\_attributes (15–16 items):} each $\le$6 words, pattern {\tt [part]: [descriptor]}; may prefix {\tt often:} for $40\text{--}60\%$ traits; cover $\le 8$ distinct parts; $\le 2$ items per part unless clearly distinct \& discriminative; order by utility—positions 1–6 hard-localizable parts (best for patching/grounding), 7–12 global appearance/pose/material, 13–16 auxiliary/negative/background (logos/wordmarks/stable background). Prefer attributes that directly answer likely QA.
\end{itemize}

\K{cVocab}{Descriptors \& vocab (preferred, open-world allowed).}
\begin{itemize}[leftmargin=4.2mm,itemsep=1.5pt,topsep=2pt]
  \item \K{cVocab}{Colors:} black, white, gray, silver, gold, red, orange, yellow, green, blue, purple, pink, brown, beige, navy, teal; concise free forms allowed (e.g., {\tt blue-green}, {\tt warm gray}).
  \item \K{cVocab}{Patterns:} solid, stripes, polka dots, plaid, floral, check, camo, gradient, logo, wordmark; concise free forms allowed.
  \item \K{cVocab}{Shapes:} round, square, rectangular, oval, triangular, tapered, curved, flat, ribbed, ridged, beveled, domed; concise free forms allowed.
  \item \K{cVocab}{Materials:} cotton, denim, leather, metal, plastic, wood, ceramic, glass, fabric, rubber, fleece, knit, plush; concise free forms allowed.
  \item \K{cVocab}{Accessories:} glasses, hat, cap, bow, scarf, helmet, tie, lanyard, headphones.
  \item \K{cVocab}{Species/breed tokens:} when visually warranted are allowed but not required (dog, cat, shiba inu, husky, corgi, ragdoll, british shorthair).
\end{itemize}

\K{cParts}{Parts vocabulary (singular, reusable).}
\begin{itemize}[leftmargin=4.2mm,itemsep=1.5pt,topsep=2pt]
  \item \K{cParts}{Core parts:} hair, ear, eye, face, beard, whiskers, muzzle, tail, collar, chest, back, belly, paw, hand, beak, horn, fin, wing, handle, lid, rim, body, keycap, roof, spire, window, door, levels, inscription, strap, pocket, sleeve, cuff, collarband, button, zipper, logo, wordmark, glasses, hat, bow, scarf, helmet, tie, lanyard, headphones, breed, species, pose, gesture, background, base.
  \item \K{cParts}{Extension rule:} If a salient part is missing, introduce a concise new singular noun and reuse it consistently.
\end{itemize}

\K{cCoverage}{Coverage guidance (use when applicable).}
\begin{itemize}[leftmargin=4.2mm,itemsep=1.5pt,topsep=2pt]
  \item \K{cCoverage}{Animals/toys:} species/breed, dominant coat colors, white/black presence, eye color, whiskers, tail presence/curvature, collar presence, material, pose, cartoon/humanlike if applicable.
  \item \K{cCoverage}{Cups/mugs:} body shape, handle presence, material, rim, lid presence, motif/face cues, dominant color, exact wordmark text when present.
  \item \K{cCoverage}{Devices:} body shape, body color, keycap color/pattern, logo/wordmark when present.
  \item \K{cCoverage}{Buildings/landmarks:} primary color/material, roof presence, door/window, spire count, levels count, style cue, inscription presence, base/platform, background.
  \item \K{cCoverage}{Persons:} glasses, hair length/texture/color, beard/bald, adult/child, top type \& color/pattern, accessories, gestures, visible wordmark/logo/number on clothing, background.
\end{itemize}

\K{cAttr}{Attribute formatting examples (illustrative).}
\begin{itemize}[leftmargin=4.2mm,itemsep=1.5pt,topsep=2pt]
  \item \K{cAttr}{Animal/toy:} {\tt species: dog}, {\tt breed: shiba inu}, {\tt tail: curled}, {\tt collar: present}, {\tt whiskers: present}, {\tt body: orange solid}, {\tt material: plush}
  \item \K{cAttr}{Cup:} {\tt body: round}, {\tt handle: present}, {\tt material: ceramic}, {\tt rim: domed}, {\tt wordmark: Neural Information Processing Systems}
  \item \K{cAttr}{Person:} {\tt glasses: present}, {\tt hair: black short}, {\tt beard: absent}, {\tt top: black solid}, {\tt logo: Microsoft}, {\tt gesture: peace-sign}, {\tt background: lake}
  \item \K{cAttr}{Building:} {\tt spire: count-2}, {\tt levels: count-7}, {\tt roof: present}, {\tt inscription: present}, {\tt base: rocks}, {\tt body: gothic}
\end{itemize}

\K{cQuality}{Quality \& count checks (silent, mandatory).}
\begin{itemize}[leftmargin=4.2mm,itemsep=1.5pt,topsep=2pt]
  \item \K{cQuality}{Output JSON only:} no prose, no code fences, no extra keys, no trailing commas.
  \item \K{cQuality}{Exactly four keys:} in the required order.
  \item \K{cQuality}{Caption length:} 24–30 words; ordering \& paraphrasing rules respected; optional palette ranking allowed.
  \item \K{cQuality}{Fingerprint size:} {\tt fingerprint\_attributes} has 15–16 items; each $\le$6 words; $\le$8 parts; $\le$2 per part unless clearly distinct \& discriminative; positions 1–6 are hard-localizable cues.
  \item \K{cQuality}{Thresholds:} majority thresholds \& tie-breaking by discriminativeness respected; exact text preserved for wordmarks/logos; lowercase otherwise.
  \item \K{cQuality}{Unknown concept:} If concept name is uncertain, set {\tt "concept": "unknown"} but still complete caption and attributes from majority-evident traits.
\end{itemize}

\K{cFinal}{Final output requirement.} Return the final JSON {\bf only}, conforming to all rules above.

\end{tcolorbox}

\paragraph{Discussion}
The instruction biases the model toward compact, discriminative, and patch-friendly attributes. The majority-threshold policy, negative evidence allowance, and strict formatting constraints jointly reduce hallucination and enforce consistent field semantics across concepts. The preprocessing and JSON normalization further ensure that downstream retrieval and grounding operate over a predictable schema with bounded size and stable key ordering.

\section{Additional Qualitative Results}
\label{app:additional-results}
In this section, we provide more qualitative results in Table~\ref{tab:details-recall}.

\begin{table}[ht]
  \centering
  \caption{More examples of qualitative results.}
  \label{tab:details-recall}

  \renewcommand{\arraystretch}{1.18}

  \begin{tabularx}{\textwidth}{>{\centering\arraybackslash}m{3.3cm} >{\centering\arraybackslash}m{\dimexpr\textwidth-3.3cm-4\tabcolsep\relax}}
    \toprule
    Concept & Conversation \\
    \midrule
    \addlinespace[6pt]
    \raisebox{-.5\height}{\begin{minipage}[c]{3.3cm}
      \centering
      \includegraphics[width=3cm]{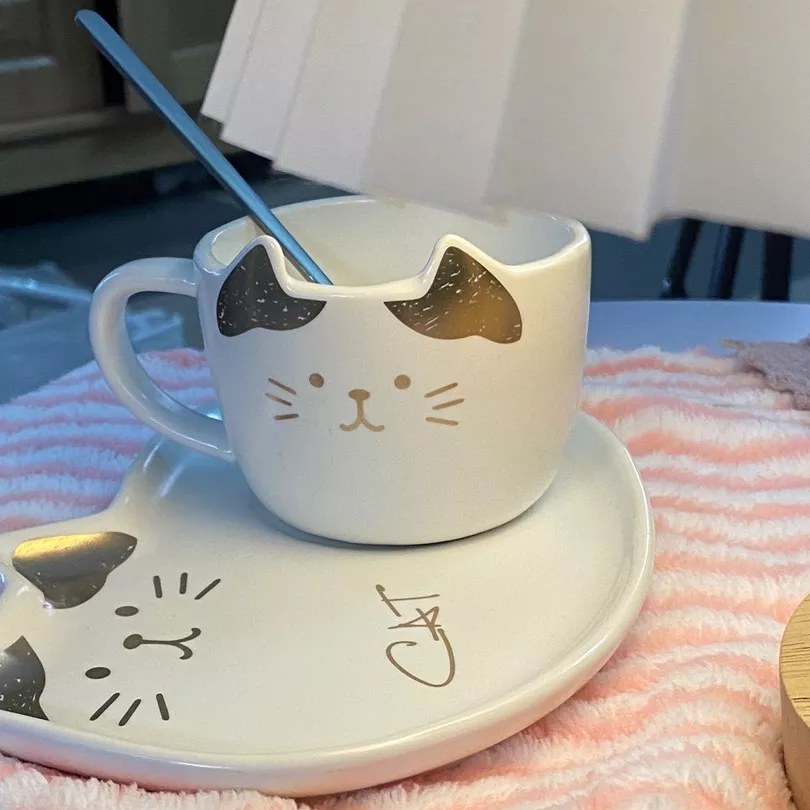}\\
      \textless cat-cup\textgreater
    \end{minipage}}
    &
    \raisebox{-.5\height}{\begin{minipage}[c]{\linewidth}
      \setlength{\baselineskip}{1.05\baselineskip}
      \raggedright
      Question: What kind of ear does \textless cat-cup\textgreater \space
      have?\\[2pt]
      \setlength{\fboxsep}{2pt}
      \colorbox{gray!20}{\makebox[\linewidth][l]{\textbf{Jarvis:} Metallic brown.}}\\[6pt]
      Question: What word is written on \textless cat-cup\textgreater's matching saucer?\\[2pt]
      \setlength{\fboxsep}{2pt}
      \colorbox{gray!20}{\makebox[\linewidth][l]{\textbf{Jarvis:} CAT.}}\\[6pt]
      Question: How would you describe \textless cat-cup\textgreater's overall shape?\\[2pt]
      \setlength{\fboxsep}{2pt}
      \colorbox{gray!20}{\makebox[\linewidth][l]{\textbf{Jarvis:} Cat Cup has a rounded mug silhouette.}}\\[6pt]
    \end{minipage}} \\
    \addlinespace[4pt]
    \midrule
    \addlinespace[6pt]
    \raisebox{-.5\height}{\begin{minipage}[c]{3.3cm}
      \centering
      \includegraphics[width=3cm]{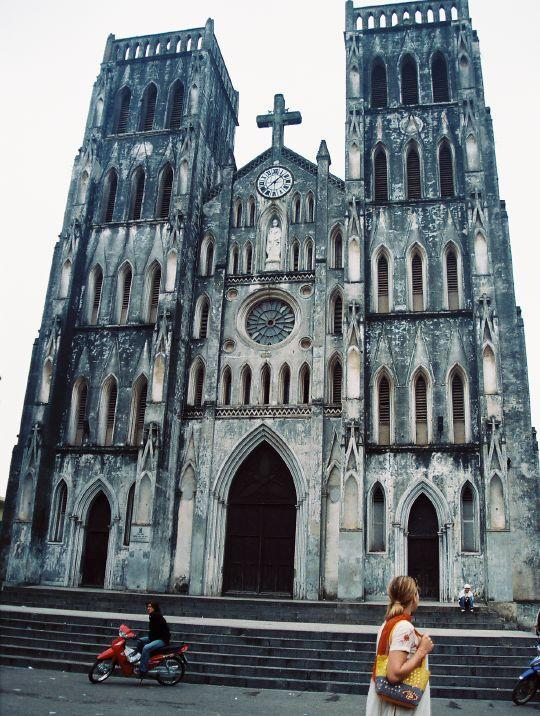}\\
      \textless nha-tho-hanoi\textgreater
    \end{minipage}}
    &
    \raisebox{-.5\height}{\begin{minipage}[c]{\linewidth}
      \setlength{\baselineskip}{1.05\baselineskip}
      \raggedright
      Question: How many floors does \textless nha-tho-hanoi\textgreater \space have?\\[2pt]
      \setlength{\fboxsep}{2pt}
      \colorbox{gray!20}{\makebox[\linewidth][l]{\textbf{Jarvis:} Nha Tho Hanoi has four floors.}}\\[6pt]
      
      Question: How would you characterize the shape of \textless nha-tho-hanoi\textgreater's roof?\\[2pt]
      \setlength{\fboxsep}{2pt}
      \colorbox{gray!20}{\makebox[\linewidth][l]{\textbf{Jarvis:} Nha Tho Hanoi's roof has a flat top.}}\\[6pt]
      
      Question: What's special about \textless nha-tho-hanoi\textgreater's architectural design?\\[2pt]
      \setlength{\fboxsep}{2pt}
      \colorbox{gray!20}{%
      \parbox{\linewidth}{%
        \textbf{Jarvis:} Nha Tho Hanoi has a tall twin-tower silhouette,     weathered gray stone facade, pointed arched windows and a large circular rose window, with a central cross and clock above the main entrance.%
        }%
      } \\[6pt]
      
    \end{minipage}} \\[6pt]
    \addlinespace[4pt]
    \midrule
    \addlinespace[6pt]
    \raisebox{-.5\height}{\begin{minipage}[c]{3.3cm}
      \centering
      \includegraphics[width=3cm]{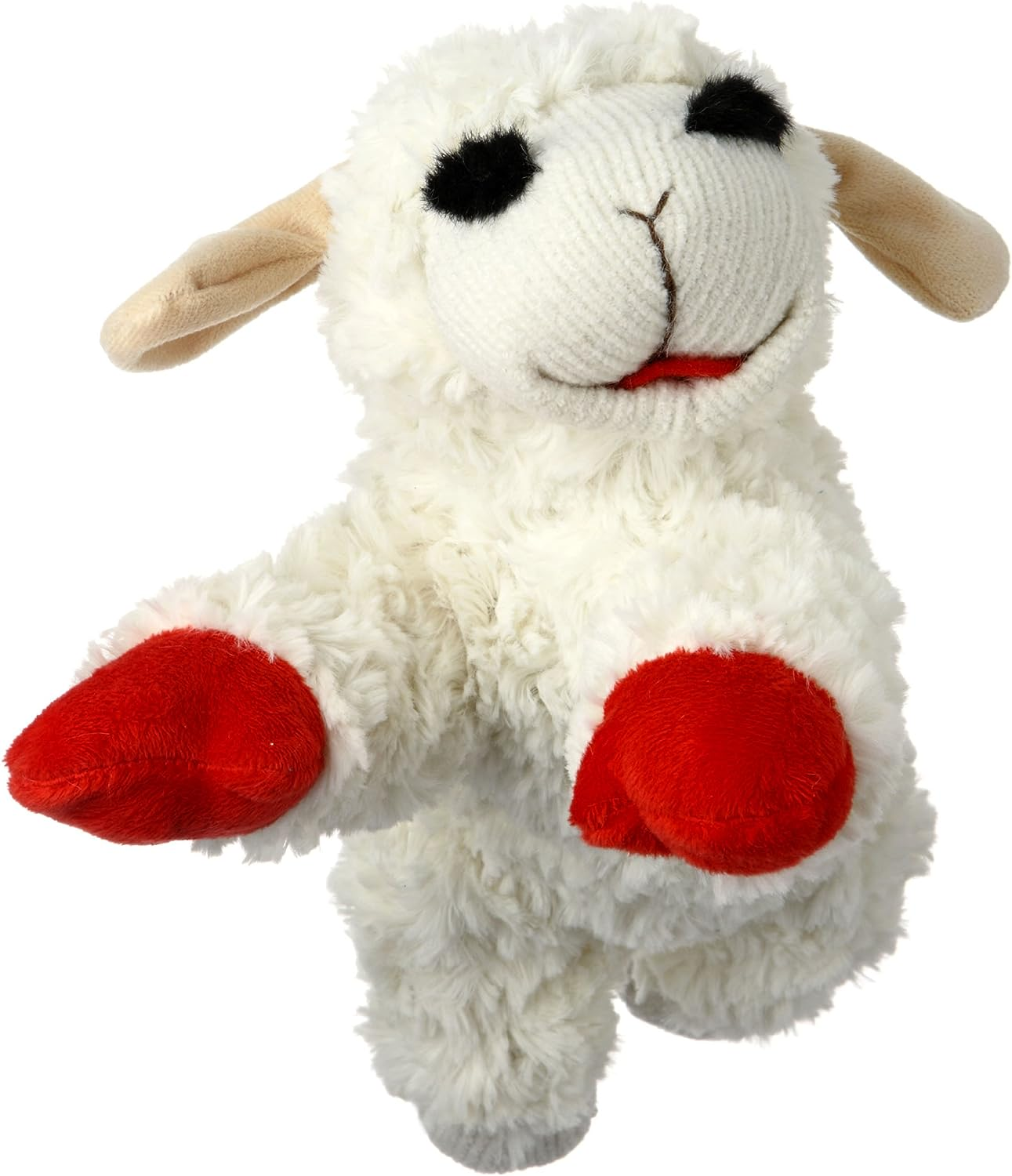}\\
      \textless lamb\textgreater
    \end{minipage}}
    &
    \raisebox{-.5\height}{\begin{minipage}[c]{\linewidth}
      \setlength{\baselineskip}{1.05\baselineskip}
      \raggedright
      Question: Describe \textless lamb\textgreater's ear.\\[2pt]
      \setlength{\fboxsep}{2pt}
      \colorbox{gray!20}{\makebox[\linewidth][l]{\textbf{Jarvis:} Lamb's ear is beige and floppy.}}\\[6pt]
      
      Question: What's special about \textless lamb\textgreater's appearance?\\[2pt]
      \setlength{\fboxsep}{2pt}
      \colorbox{gray!20}{%
      \parbox{\linewidth}{%
        \textbf{Jarvis:} Lamb has a white fleece texture, beige ears, and bright red hooves.%
        }%
      } \\[6pt]
      
      Question: Does \textless lamb\textgreater have any accessory attached?\\[2pt]
      \setlength{\fboxsep}{2pt}
      \colorbox{gray!20}{%
      \parbox{\linewidth}{%
        \textbf{Jarvis:} No, Lamb does not have any accessory attached.%
        }%
      } \\[6pt]
      \end{minipage}} \\[6pt]
    \bottomrule
  \end{tabularx}
\end{table}

\end{document}